\documentclass[10pt,twocolumn,letterpaper]{article}

\usepackage{iccv}
\usepackage{times}
\usepackage{epsfig}
\usepackage{graphicx}
\usepackage{amsmath}
\usepackage{amssymb}
\usepackage[switch]{lineno} 
\usepackage{subcaption}
\usepackage{amsmath}
\usepackage{amsfonts}
\usepackage{algorithmicx,algorithm}
\usepackage[noend]{algpseudocode}
\usepackage{booktabs}
\usepackage{multirow}
\usepackage{makecell}
\usepackage[inline]{enumitem}
	
\usepackage{color, colortbl}

\newcommand{\act}{a_t}
\newcommand{\obs}{o_t}

\algnewcommand{\LineComment}[1]{\State \(\triangleright\) #1}
\let\sss= \scriptscriptstyle
\graphicspath{{fig/}}

\definecolor{Gray}{gray}{0.9}

\usepackage[pagebackref=true,breaklinks=true,letterpaper=true,colorlinks,bookmarks=false]{hyperref}

\iccvfinalcopy 

\def\iccvPaperID{9376} 

\ificcvfinal\fi

\begin{document}

\title{What to Do Next?\\Memorizing skills from Egocentric Instructional Video}

\author{Jing Bi\\University of Rochester\\\texttt{jingbi@cs.rochester.edu}
\and
Chenliang Xu\\University of Rochester\\\texttt{chenliang.xu@rochester.edu}
}

\maketitle
\thispagestyle{empty}

\begin{abstract}
Learning to perform activities through demonstration requires extracting meaningful information about the environment from observations. 
In this research, we investigate the challenge of planning high-level goal-oriented actions in a simulation setting from an egocentric perspective. 
We present a novel task, interactive action planning, and propose an approach that combines topological affordance memory with transformer architecture. 
The process of memorizing the environment's structure through extracting affordances facilitates selecting appropriate actions based on the context. 
Moreover, the memory model allows us to detect action deviations while accomplishing specific objectives. 
To assess the method's versatility, we evaluate it in a realistic interactive simulation environment. 
Our experimental results demonstrate that the proposed approach learns meaningful representations, resulting in improved performance and robust when action deviations occur.
\end{abstract}
\vspace{-5mm}

\section{Introduction}

Recent advances in the field of egocentric vision have enabled the recognition and understanding of human activities in real-world scenarios, paving the way for a range of applications, such as augmented reality assistants that guide users in performing tasks  and human-robot collaboration systems. 
 Significant progress has been made in this field with the focus on recognizing \textit{What is happening?}~\cite{zhu2020comprehensive}, 
\eg action recognition~\cite{duan2022revisiting} and detection~\cite{Zhao_2022_CVPR,Dai_2022_CVPR,Ou_2022_CVPR}.
 
\begin{figure*}[ht!]
\centering
    \includegraphics[width=0.95\textwidth]{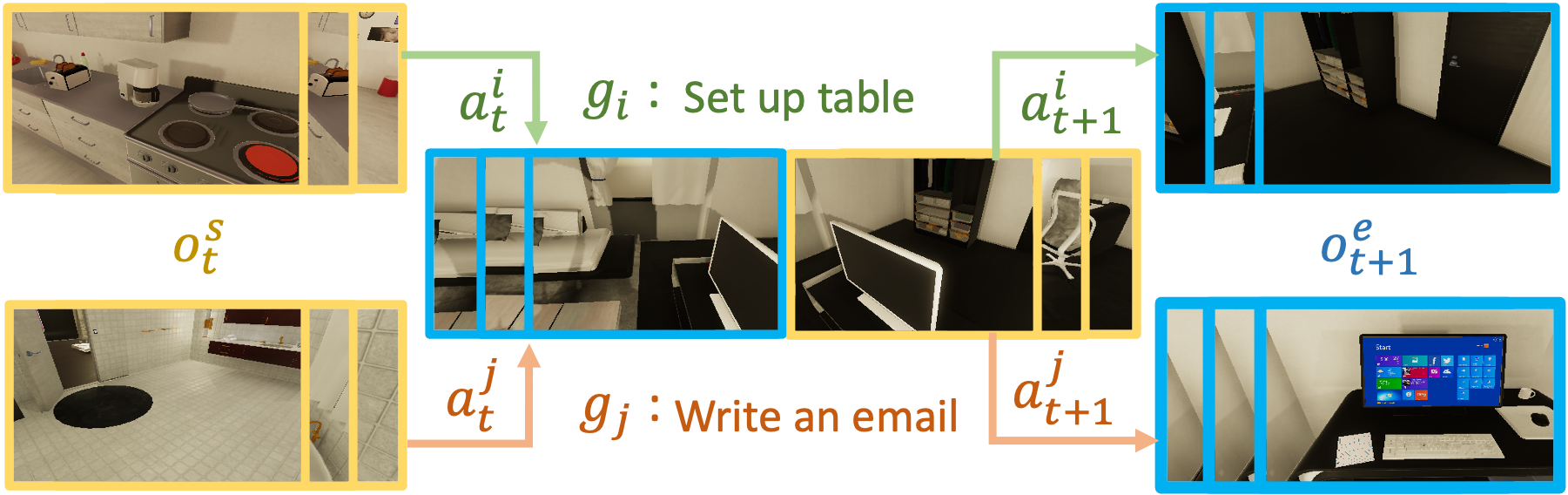}
    \caption{\textbf{The problem formulation:} Given the description of the goal $g_i$ \textit{set up table} and initial observation, the agent should need to plan a sequence of actions to achieve the task. When executing an action $a^i_t$, the agent will receive a stack of $[o^s_t, o^e_t]$ as the observation, where $o^s_t, o^e_t$ represent the starting/ending point of the action. Different trajectories may share the same scene; thus, understanding the environment is crucial for action planning}
    \label{fig:case}
    \vspace{-4mm}
\end{figure*}

However, visual intelligence demands more than mere comprehension of current events but also the capacity to achieve specific task with perceived information.
This generally involves two steps. 1) Task planning: plan ahead a series of steps (high-level task actions, \eg, \textit{navigate to kitchen}). 2) Motion planning: accomplish step with a sequence of motions (low-level actions, \eg, \textit{turn right, turn left}) based on the objectives and the state of the environment.
Recent advances in visual navigation~\cite{anderson2018vision,kolve2017ai2} go beyond recognition and equip autonomous agents with ability to understand the sorrounding environment and make decisions with visual information which is crucial for solving \textbf{how to} move to accomplish given step. 
In this study, we focus on the visual task planning: deciding \textbf{what to do next} to achieve the given task with only perceived visual information.
As shown in Fig.\ref{fig:case}, with a description of the task and the visual observation of the initial location, a successful involves first navigate to kitchen, grab plate and place them on the table.

The classic task planning problem has been extensively studied in the field of robotics, but it assumes a closed world where the algorithm is equipped with complete domain knowledge. This means that all possible states and actions are explicitly defined in logic form using description languages like Planning Domain Description Language (PDDL) ~\cite{aeronautiques1998pddl,fox2003pddl2}. 
For instance, in a block world, the state of the world is defined by a set of predicates such as \texttt{(:init (on block-a table) (on block-b block-a))} and \texttt{(:goal (on block-b table) (on block-a block-b))}. 
The planner then uses a search algorithm to determine the minimum number of actions required to transform the current state into the goal state. 
However, this method is not suitable for visual inputs, as representing and grounding visual observations in a symbolic form is still a challenging task ~\cite{gupta2017cognitive,lecun2015deep}.

A similar work, Visual semantic planning~\cite{zhu2017visual}, also focuses on task planning and involves predicting task actions in a simulation environment using pre-defined high-level actions. 
Their method utilizes Q-learning with successor feature vectors to encode the expected outcomes of future actions and reason about the optimal action based on future rewards.
However, their approach relies on optimal planning and modeling reward functions, while our method does not have such constraints.
A similar problem, referred to as procedure planning in instructional videos, has recently been proposed~\cite{chang2019procedure,bi2021procedure} where the observation is clip of video and the algorithm is required to plan a sequence of actions.
However, evaluating the planning model's performance with a static dataset often produces overoptimistic results since a feedback signal is only provided if the planned actions are correct. 
For example, in a breakfast preparation task, the correct step may be to \textit{cut an apple} but if the system makes an incorrect prediction such as \textit{pick an apple} the dataset will still provide a frame of cut apples instead of a complete one. 
The lack of reactive feedback limits the ability to predict suitable future actions and makes it difficult to identify and recover from such errors. 

To address the aforementioned issues, we propose a novel task called Interactive Action Planning that emphasizes the following key elements:
\begin{enumerate*}[label=\alph*)]
    \item learning skills offline from recorded experience without the need for interaction;
    \item online evaluation of performance with an interactive environment;
    \item comprehending the description of the given task;
    \item planning a sequence of actions while being aware of the environment, as the one shown in  Fig.~\ref{fig:case}.
\end{enumerate*}

The proposed task aims to overcome the challenges posed by the limitations of current visual task planning problem, and offers a more comprehensive and effective evaluation pipeline. 
Unlike typical image-language translation tasks, predicting the same actions as ground truth with sequence mapping is not optimal because the same task can be executed differently based on the situation.

Solving the proposed task requires the agent to reason about the objectives of the activities,  determine the possible interactions in the current situation, and decide the appropriate actions~\cite{casas2018intentnet}. 
As humans, it is difficult to approach new problems as if we have never faced them before.
Instead, we try to find the best plan they have heard of or previously used that is closest to the problem at hand and attempt to adapt it to the current situation.
Similarly, when planning in a new environment, it is crucial for the agent to extract useful information from past experiences.
In light of classic case-based planning approaches that view planning as a memory task, and recent work of utilizing episodic memory for control~\cite{pritzel2017neural}, we tackle this problem with focusing on two aspects: memorizing essential information from demonstrations with awareness of the environments and retrieve goal-directed  relevant knowledge for planning.
We proposed a single holistic framework that integrates a new form of memory to associate the visual information with environment affordance.
Specifically, the proposed memory serves a few purposes:
\begin{enumerate*}[label=\alph*)]
    \item organizing the expert's experience into a more structured representation for subsequent action policy learning;
    \item localizing current visual observations in the latent space and retrieving corresponding memories when encountering a similar situation to narrow down potential action selection
    \item enabling replanning once the action execution deviates from reaching the goal.
\end{enumerate*}
This framework has the advantage of decomposing the action learning from memorizing the environment, enhancing the agent's ability to retrieve and apply past experiences to improve planning in novel situations.

We developed several new evaluation methods to compare the effectiveness of our approach in both static and interactive scenarios. 
Our results demonstrate the robustness of our planning algorithm, even in the presence of action deviation. This confirms that our memory learns meaningful representations of the environment, which is useful for efficient planning in the latent space.

The main contributions of our work are summarized as follows:
\begin{enumerate*}[label=\alph*)]
\item proposing a novel task called interactive action planning that emphasizes more realistic learning settings
\item we propose a new form of memory that learns to associate the demonstration from different episodes and embeds affordance information to convey the current situation that helps subsequent action generation.
\item proposing a replan algorithm based on adversarial attack with localization network to find an alternative valid plan when action deviation occurs
\item we further introduce several practical evaluation methods to complement existing evaluation metrics.
\end{enumerate*}

\begin{figure*}[t!]
\centering
    \includegraphics[width=\textwidth]{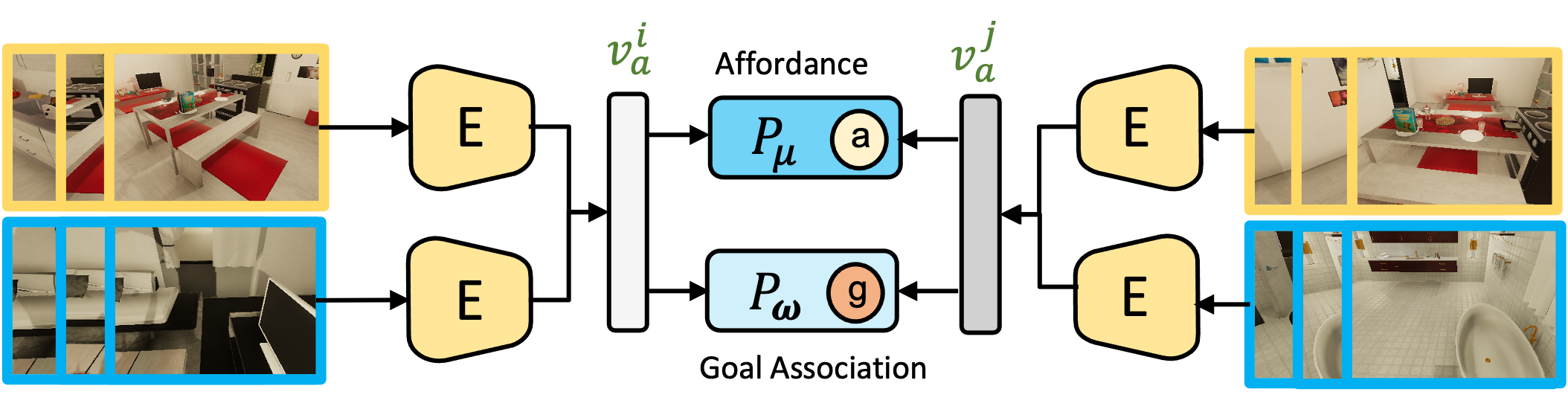}
	\caption{\textbf{Topological Affordance Memory:} We embed the visual observation $[o^s_t, o^e_t]$ as $v^i_a,v^j_a$ representing the affordance  $i,j$. The two encoded feature will be projected to latent space for further contrastive learning. The goal association function conditioned on given goal $g$ and tries to discriminate whether two features are associated with $g$. The affordance function learns to assign high value for pair of features with same actions and low scores otherwise}
    \label{fig:mem}
        \vspace{-3mm}
\end{figure*}


\section{Related work}
\subsection{Learning actions with visual perception}
In recent years, there has been significant research interest in learning action policies from visual information through interactions with simulation environments, such as visual navigation and embodied question answering, as explored by works like ~\cite{zhu2017target,dwivedi2022navigation,kolve2017ai2,anderson2018vision,das2018embodied}. 
Often learning with a well-defined reward function with reinforcement learning, these work focus on how to achieve the given goal with low-level motions, which servers well as a subroutine of the high-level planned action.

Another line of work is trying to learn action policy from recorded video with Imitation learning
Although the static dataset cannot provide such a signal, a straightforward formalization of the reward can be defined as a binary signal: when the task is completed, we mark it as a success or failure.
However, when the task requires multiple steps to finish, this formalization often suffers from the sparse reward problem as the agent receives little or no feedback.

Pritzel \etal~\cite{pritzel2017neural} proposed an external episodic memory to store experience to tackle this limitation. 
This memory module used state representation as the key and the returned future reward as a value to construct the memory that can be used to query the Q-value during training.
In contrast, our work focuses on constructing memory from human demonstrations without the need for rewards, requiring a comprehensive understanding of the environment ~\cite{levine2020offline}.

\subsection{Memory for planning}
Recent research has incorporated graph-based memory with classic planning algorithms by embedding observations as graph nodes and using a learned distance metric to compute edges ~\cite{SavinovDosovitskiyKoltun2018_SPTM}. This allows for the shortest-path algorithm (Dijkstra) to be applied to the learned latent space, where the agent can retrieve suitable actions from memory. 

Variations of this approach include Search on the Replay Buffer ~\cite{eysenbach2019search}, which combines the strengths of planning and reinforcement learning by using graph search over the replay buffer, and Sparse Graphical Memory ~\cite{emmons2020sparse}, which consolidates many environment states and sparsifies the dense graph for improved planning performance. 
Another related method is Hallucinative Topological Memory (HTM) ~\cite{liu2020hallucinative}, which employs hallucination techniques to explore possible future states of the environment and utilizes a contrastive energy function to construct more accurate edges between memory nodes.

However, unlike these previous works that focus on low-level control tasks such as navigation and manipulation, our approach aims to plan high-level action primitives from human demonstrations.

\subsection{Contrastive Learning}
Contrastive Learning (CL) has emerged as a key component in self-supervised learning for computer vision, aiming to learn an embedding space that preserves similarity relationships between samples. Triplet loss, initially proposed in FaceNet~\cite{schroff2015facenet}, has been used to learn representations of human faces with different poses and angles. The InfoNCE loss in CPC~\cite{oord2018representation} extends the original NCE loss and employs categorical cross-entropy loss to differentiate positive samples from a set of unrelated noise samples.

While most variants of CL operate in an unsupervised setting, we employ supervised CL~\cite{NEURIPS2020_d89a66c7}, which extends the InfoNCE loss by incorporating label information. Positive samples are generated by augmenting anchoring examples with transformation functions, while we consider the augmented versions of human activities as the same action but in different situations. Our approach adapts the supervised CL loss to our multi-label video classification problem, which differs from previous work that focuses on single-label image classification.


\section{Method}

In contrast to previous work ~\cite{liao2019synthesizing} using simulator internal scene graphs that captures full state of the environment, we focus on using only egocentric images as observations that provides only partial information about the surroundings.
This leads us to  formalized the given task as partially observable Markov decision processes(POMDPs) defined as $(\mathcal{O},\mathcal{S},\mathcal{A},\mathcal{T},\mathcal{R},\mathcal{C})$.

As shown in Fig.~\ref{fig:case}, imaging an agent is standing in a washroom, and the egocentric observation $o_t^s$ only captures the portion of the environment that is directly visible to the agent.
Now suppose the agent needs to complete the task of setting up table, which involves navigating to different locations in the room.
The agent needs to be informed not just by the current observation but also by its past experience and the actions taken so far.

We adopt a standard imitation learning (IL) setup where the agent learns the action policy from demonstrations. 
Specifically, we have access to a set of egocentric instructional video of an expert performing a given task ${(o_{\sss{1:T}}^j,a_{\sss{1:T}}^j), g}{j=0}^{\sss K}{\sim}\pi{E}$.
Moreover, when executing a high-level action, it may take an arbitrary amount of time to complete. 
In our approach, we only choose the starting $f_k$ frames as $o^s_t$ and the same number of frames at the end as $o^e_t$, where $f_k=5$ in our experiments, as shown in Fig.~\ref{fig:case}. 
At each time step $t$, the agent receives a stack of frames $[o^s_t, o^e_t]$ as the observation $\obs$ of the surroundings and selects an action $\act$ from a set of actions $A$ such as \textit{Moving to Kitchen}.
Notably, the set of actions in $A$ may vary depending on the agent's current location, for example, the agent cannot \textit{pickup apple} if there is no apple in its view. 

Our insight is that a successful plan relies on answering the following two questions:
\begin{enumerate*}[label=\alph*)]
    \item What actions are possible in the current situation (environment affordances)? 
    \item What actions need to be performed to achieve the desired goal (goal-directed actions)?
\end{enumerate*}
To this end, we propose a framework consisting of two main components:
\begin{enumerate*}[label=\alph*)]
	\item Topological Affordance Memory: We construct a structured memory that associates environment affordances with visual information from different episodes of experience.
	\item Action Generation module: We utilize the learned memory graph to generate a valid sequence of actions $a_{\sss{1:T}}$ to achieve the indicated goal. This action generation process includes three sub-steps: localization, memory retrieval, and planning. By breaking down the task in this way, our approach enables efficient and effective planning and execution of complex tasks in dynamic environments, as illustrated in Fig.~\ref{fig:plan}.
\end{enumerate*}
\begin{figure*}[t!]
\centering
    \includegraphics[width=0.85\textwidth]{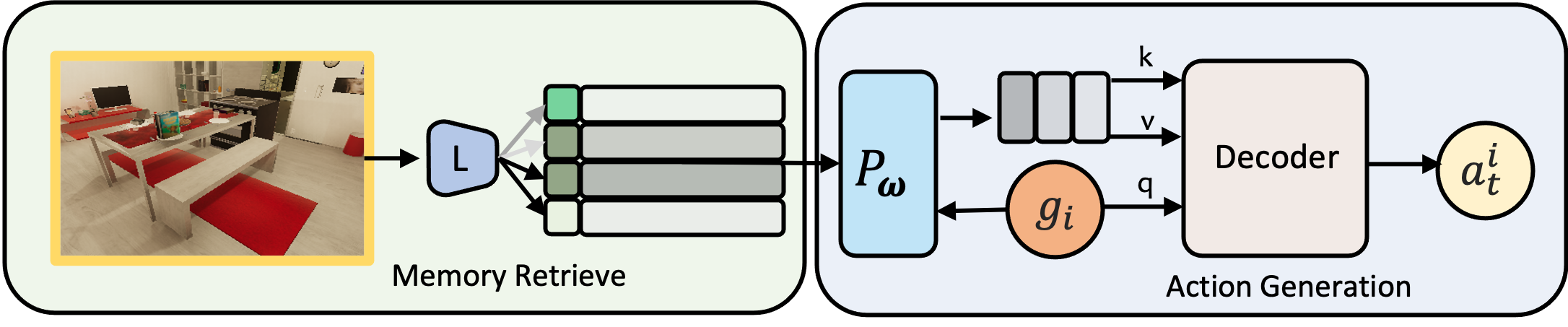}
	\caption{\textbf{The Planning Module:} Starting from the initial observation frame, we first use the localization step to identify the current situation in the learned memory. The darkness of the arrow in the figure represents the similarity between the frame representation vector and the key of the memory. Once we retrieve a similar situation from the memory, we use the goal association function $P_\mu$ to check whether the current progress aligns with the indicated goal. If necessary, we initiate replanning. Finally, we feed the goal, action history, and retrieved memory to the Transformer decoder to plan the next action.}
    \label{fig:plan}
    \vspace{-3mm}
\end{figure*}
  

\subsection{Topological Affordance Memory}

Research in psychology and cognitive neuroscience has shown that the hippocampus plays a crucial role in human memory by associating past experiences and remembering their relationships~\cite{nelson1995ontogeny, anderson2014human, kohonen2012associative}.
This motivates us to propose a new memory structure called Topological Affordance Memory~(TAM), which stores the expert's past successful experiences in achieving specific goals.

It consists of three components:
\begin{enumerate*}[label=\alph*)]
    \item Localization $L$: The localization network identifies the most similar node in the graph based on the current visual observation for memory retrieval
    \item Affordance Learning $\mathcal{P}\mu$: This module learns to capture the information about what actions can be taken in the current situation.
    \item Goal association $\mathcal{P}\sigma$: Given that there may be multiple ways to achieve a given goal, this module is used in planning to verify whether the agent has deviated from the current goal.
\end{enumerate*}

The key and value of TAM's nodes are both learned representations, where the representations learned through $L$ serve as the key for each node in the graph and the values is learned by  $\mathcal{P}\mu$,  $\mathcal{P}\sigma$.
we will discuss how to construct memory with these components as following and how to leverage them for planning in Sec.\ref{sec:act}.

\noindent \textbf{Affordance Learning}
To effectively learn the affordances of the environment, it is crucial for the agent to associate similar scenes that can afford the same interaction.
We tackle this as a supervised contrastive learning task, leveraging the availability of action labels for each observation.
In contrast of typical CL using randomly transformed images as different views of the original data, we take a different approach. 
We define augmented versions of observation as the ones that share the same action from other episodes across the entire demonstration.
This allows us to compare and contrast similar scenes that afford similar actions, while disregarding scenes that are not relevant.
We employ the R2+1D18~\cite{tran2018closer} video encoder to embed the observation into a representation vector, denoted as $v^i_a=Enc(\obs)\in R^D$. 
For affordance projection function $\mathcal{P}_\mu(\cdot)$, we instantiate it as a 3-layer multi-layer perceptron(MLP)  $z_i = \mathcal{P}_\mu(v^i_a)$ and we adopt the InfoNCE~\cite{van2018representation} loss with temperature $\tau$ as our affordance loss $\mathcal{L}_{\text {\rm aff }}$ is defined as:
\begin{equation}
    \mathcal{L}_{\text {\rm aff }}=\sum_{i \in I} \frac{-1}{|Pos(i)|} \sum_{p \in Pos(i)} \log \frac{\exp \left(z_{i} \cdot z_{p} / \tau\right)}{\sum_{z_{a} \in A(i)} \exp \left(z_{i} \cdot z_{a} / \tau\right)},
\end{equation}
where $Pos(i)$ represents the positive set of projected feature $z_i$ and $|P(i)|$ represents its cardinality.

\noindent \textbf{Goal Association}
Our goal association function $\mathcal{P}_\sigma(v^i_a,v^j_a,g_i)$ shares the same encoder with $\mathcal{P}_\mu(\cdot)$ but serves as a discriminator to identify if two $(v^i_a,v^j_a)$ node are attempting to reach the same goal, which is trained with Binary Cross Entropy loss. 
The goal association function $\mathcal{P}_\sigma(v^i_a,v^j_a,g_i)$ will be used in the action generation step.

\noindent \textbf{Localization and Retrieve}
The above two modules focused on learning representation for node value, whereas the role of $L$ is to estimate the similarity between two frames $(f_1, f_2)$ which is then used as a key for the corresponding node. 
The objective of $L$ is to assign high scores to pairs of temporally adjacent frames and low scores to temporally distant pairs, allowing us to frame this as a self-supervised task in which $L$ predicts the probability of the given two frames being temporally close.
To achieve this, we use training triples $(f_i, f_j, y)$ consisting of two observed frames and a binary label indicating their temporal closeness. 
Positive samples are pairs inside consecutive video clips $(f_i,f_j)\in [o^e_{i-1},o^s_i]$, while negative samples are separated apart. 
We utilize a Siamese architecture with ResNet18~\cite{he2016deep} as the image encoder for $L$
This approach allows us to store and retrieval past successful experiences when encountering similar situations.


\begin{algorithm}[t]
	\hspace*{0.02in} {\bf Input:}
	TAM memory $M_{T}$, observed frame $f_t$, image encoder $E$, localization network $L$, goal association network $\mathcal{P}_\mu(\cdot)$, threshold $T$, trial time $K$, last step localized node $n_{t-1}$, objective function $\rm Loss$

	\begin{algorithmic}[1]
		\caption{Replan}
		\label{alg:replan}
		\LineComment{Find the nearest node}
		\State $n_t = \arg\min_{n_i \in \mathcal{M}} L(E(f_t), n_i)$
		\State trail $\gets$  0
		\While{$\mathcal{P}_\mu(n_t,n_{t-1},g)< T$}
			\State $n^{k+1}_t \gets clip(n^k-\alpha \rm sign(\nabla_{n_t} Loss))$
			\State trail $\gets$ trail + 1
			\If{trail $> K$ }
			\State Break, we cannot replan
			\EndIf
		\EndWhile
		\State $n_t = n^{k}_t$
  
	\end{algorithmic}

\end{algorithm}

\subsection{Action Generation}\label{sec:act}

To improve the coherence of action sequences, we employ auto-regressive sequence models that generate actions based on the goal, action history, and retrieved memories. The model is trained to fit a distribution over an action sequence $\textbf{a} = {a_1,\dots,a_n}$ using the chain rule $p(\textbf{a})=p(a_1)\prod^T_{t=2}p(a_i|a_1,\dots,a_{i-1})$. Each term on the right-hand side is parameterized by a transformer decoder~\cite{vaswani2017attention}.

As shown in Fig.~\ref{fig:plan},the self-attention operation first processes the goal embedding and last predicted action. 
We use greedy decoding to obtain the predicted actions and add cosine positional embeddings based on the step index, while the goal positional embedding remains fixed. 

At the planning stage, we use TAM to provide a useful memory mechanism to the action generator network, as illustrated in Fig.~\ref{fig:plan}. Our approach consists of two primary steps: memory retrieval and action generation.
We encode the observation frame as a latent vector and employ the learned similarity function $L$ to retrieve K nearest neighbors from the memory, and the cross-attention module attends to the retrieved memory to predict the next action. 
This approach allows the agent to generate more coherent action sequences by integrating past experiences, the task goal and action history into the decision-making process

However, action generation often suffer from deviation when executed actions deviate from the planned actions.
To address this issue, we propose a replanning algorithm shown in Algorithm~\ref{alg:replan}, which is similar to how word processors auto-correct a sentence. 
We formulate this process as an optimization problem with the goal of maximizing the goal association score while simultaneously minimizing a modified cost function concerning the localized node. 
To accomplish this, we utilize the iterative target class method ~\cite{kurakin2018adversarial}, which repeatedly modifies the feature towards the desired loss until the node feature is aligned with the last step node feature. 
By doing so, we can ensure that even if deviations occur during execution, we can still retrieve good memory for subsequent action generation, mitigating the potential for catastrophic failure.


\section{Experiment}

In this section, we evaluate our proposed method and a baseline model in various scenarios using the Virtual home~\cite{puig2018virtualhome} environment, which provides photorealistic rendering of indoor scenes and allows interaction with 3D objects using high-level actions, such as \textit{write an email} or \textit{set up table}, as shown in Fig.\ref{fig:case}. 
All tasks are executed differently depending on the spawn locations and random preconditions. 
For instance, the step \textit{[WALK] living room} will be ignored if the spawn location is the living room. Therefore, we randomly spawned the character into four locations for each program and collected the videos.
We use the recorded video for different activities as training data and generated 10 unseen apartment for the test set to evaluate the generalizability of our proposed method.

\begin{figure}[t]
    \centering
    \includegraphics[width=\linewidth]{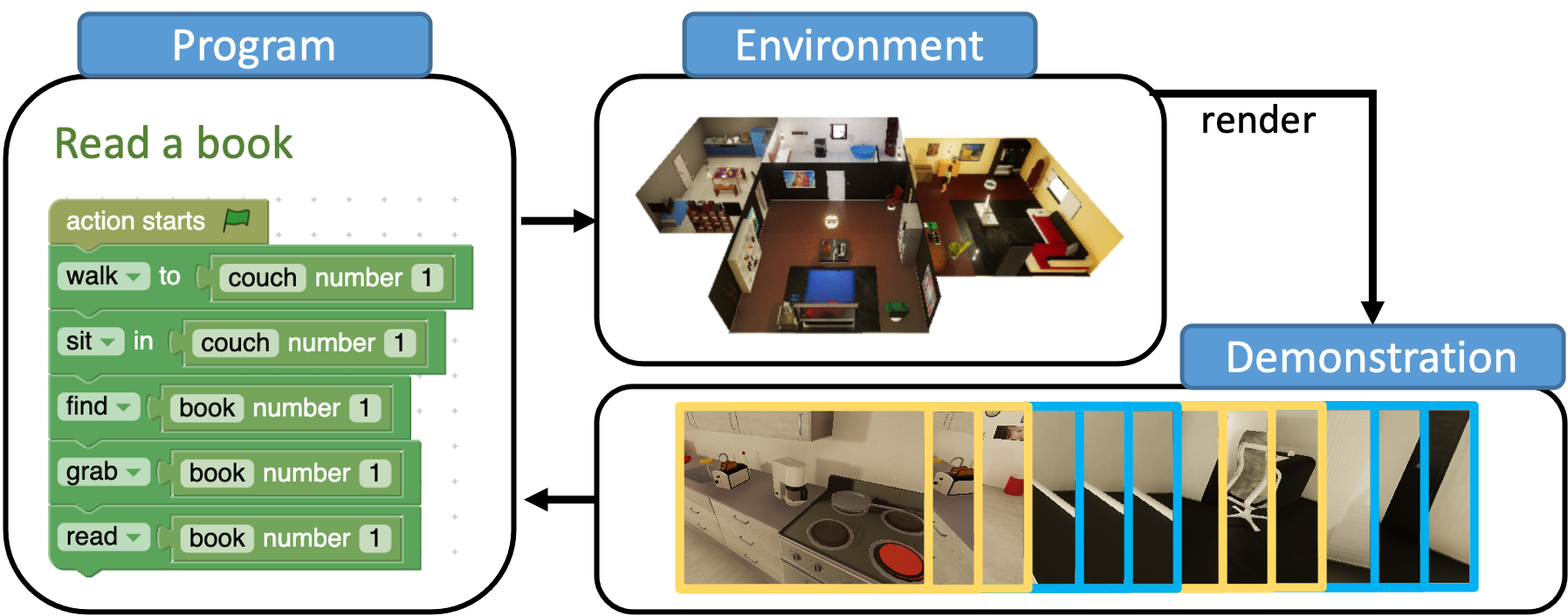}
    \caption{\textbf{Dataset collection:} Each task is formed as a sequence of high-level actions. We executed these actions in a simulated environment with built-in characters, capturing a video that serves as the expert demonstration. During evaluation, we randomly spawn the character in feasible locations to achieve the goal and begin executing the generated actions to complete the task.}
    \label{fig:data}
    \vspace{-5mm}
\end{figure}
    
\subsection{Evaluation Methods}

\noindent \textbf{Pure-text:}
Generating a sequence of actions is based solely on language descriptions without relying on visual cues.

\noindent \textbf{Visual static:}
To evaluate whether visual observation provides crucial information about the environment, we use the static dataset and provide the model with the next frame $o_{t+1}$ from the dataset, regardless of the generated action.

\noindent \textbf{Visual interactive:}
Evaluating a planning algorithm with a static dataset has limitations as it cannot provide feedback or transferred states when an unachievable action is predicted, which requires the model to understand the context of what has been done.

\noindent \textbf{Visual interactive attack:}
This evaluation setting is the most stringent scenario in whic we also want the model to be robust to deviations. 
In this setting we will randomly permute the predicted action, and the model will receive the observation from the permuted action instead of the predicted one, which requires understanding what actions are needed to achieve the given goal.
\begin{table*}[]
	\small
	\caption{\textbf{Results of Action Planning.} Our models signiﬁcantly outperform the baselines by $\sim10\%$ improvement in terms of the LCS and Executability. Our F1-score of the final graph is the most consistent with the one from executing ground-truth program. For all metrics, higher score indicates better performance. We highlighted the best performance under the most strict evaluation method. }
	\begin{tabular*}{\linewidth}{@{\extracolsep{\fill}}r|r|ccccc}
		\toprule
		~&  Methods               & LCS$\uparrow$  & Executability$\uparrow$  &F1$\uparrow$  & F1-state$\uparrow$ & F1-Relation$\uparrow$ \\ \midrule
\multirow{2}{*}{ Pure text}         & Transformer  	& 53\% &   0.13  &  0.074&  0.076 &  0.087 \\
	~	                         & ResFormer     & 69\% &   0.22  &  0.124& 	0.139 &  0.12      \\\midrule
\multirow{4}{*}{ Vis-static }        & DDN           & 52\% &   0.45    &  0.38&	0.20 &  0.25      \\
		                        & Ext-Mgail     & 77\% &   0.37  &  0.52 &  0.47  &  0.53      \\ 
		                          ~& ResFormer      & 21\% &   1     &  0.026&	0.018 &  0.028      \\
		                      ~& TAMFormer     & 81\% &   0.49  &  0.51 &  0.52  &  0.46      \\\midrule
\multirow{4}{*}{ Vis-Inter}      & DDN           & 22\% &   1     &  0.09       &	0.11 &  0.12      \\
		                      ~& Ext-Mgail     & 31\% &   1     &  0.27       &  0.29  &   0.29    \\
		                        ~& ResFormer     & 11\% &   1     &  0.01       &  0.10  &   0.10    \\
		                      ~& TAMFormer     &  \textbf{72\%} &   1 &   \textbf{0.44} &   \textbf{0.48}  &    \textbf{0.42}      \\\midrule
\multirow{3}{*}{ Vis-Inter-Att} & DDN           & 5\% &   1     &  0.09&	0.08 &  0.10      \\
		                      ~& Ext-Mgail     & 39\% &   1     &  0.23 &  0.29  &   0.24     \\
		                      ~& TAMFormer     & \textbf{64\%} &   1 &  \textbf{0.40} &  \textbf{0.42}  &   \textbf{0.39}     \\
		\bottomrule
	\end{tabular*}
	\vspace{-3mm}
	\label{table:PP}
\end{table*}
\subsection{Evaluation Metrics}

\noindent \textbf{LCS}
The normalized longest common sub-sequence(LCS) is a common measure of sequence similarity. 
We use LCS to compare the predicted and ground-truth actions for pure-text and vis-static evaluation, reporting the average overall evaluation dataset. 
For interactive evaluation, we only considered the actions that were successfully executed in the environment and compared them with the ground truth.

\noindent \textbf{Executability}
For non-interactive evaluation, we checked the executability of the generated action sequences, i.e., whether the simulator could execute them correctly given the current situation. 
Semantic correctness was the main criterion here: for example, the instruction to \textit{grab object} was only valid if the object was close enough. 
Since we only considered successfully executed actions in interactive evaluation, their executability was always true.

\noindent \textbf{Graph F1}
We defined goal completion as the change of the underlying environment states, and computed the differences between the final graphs changed by predicated actions and ground truth using F1 scores which donates as F1 in Table~\ref{table:PP}. 
Additionally, we computed the F1 scores of the object states and their relations with other objects, denoted as F1-state and F1-relation, respectively.

\subsection{Baseline}
\noindent \textbf{Transformer}~\cite{vaswani2017attention}. The Transformer is a natural choice when modeling the sequence, especially for language modeling in the pure-text evaluation. We use the Transformer as a goal encoder and action decoder for the baseline model.

\noindent \textbf{DDN}~\cite{chang2019procedure}. This method separates environment modeling and action generation. Unlike affordance learning, it captures the underlying transition information of the environment with the dynamic model and combines the classic planning algorithm and the learned model.

\noindent \textbf{Ext-Mgail}~\cite{bi2021procedure}. Like our memory structure, Ext-Mgail learns environment transition dynamic model and plannable representations upon which action sequence can generate stochastically to reach the goal. This method can be seen as a combination of goal-conditional IL and model-based IL.

\noindent \textbf{Resnet + Transformer(ResFormer)} To evaluate the importance of the visual cue, we combine additional image encoder with the vanilla Transformer. 
The first frame of each observation is encoded as an embedding vector which will stack with goal embedding together servers as the memory for the decoder. 
We use an input mask that only allows the encoder to attend the history visual embeddings.

\subsection{Results and Discussion}

Table~\ref{table:PP}  indicates that the transformer can generate plausible sequences of actions without any knowledge of the environment, as shown by the reasonably high LCS score achieved in the pure-text setting. 
However, the executability score is relatively low, indicating that the vanilla seq2seq model struggles to plan actions based on the environment.

When we added visual cues to the seq2seq model, we observed a notable improvement in the LCS score. However, the executability performance still lagged behind, indicating that the model had difficulty planning actions based on the surrounding environment. 
We suspect that the visual information enabled the model to recognize the actions better, leading to more aligned action outputs with the dataset observation. 
This suggests that visual cues provide valuable information that can improve the model's performance in planning tasks.

During interactive evaluation, the model's performance drops significantly due to changes in observation because the model has not been trained to handle variations in the environment. 
Supervised learning of action sequences may focus excessively on generating plausible action sequences and may not adequately consider the environment's current state.
Thus, when a mistake occurs, the model may not know how to recover, leading to catastrophic failures. 
This problem is particularly challenging when the model is required to plan sequences of actions to achieve a specific goal, as errors in the plan can quickly lead to failure in achieving the goal. 
Therefore, it is essential to develop models that can handle variations in the environment and can recover from mistakes during interactive evaluation.

Our proposed TAMFormer model addresses this issue by learning a memory structure that stores successful experiences, allowing it to rapidly recall potential action consequences when encountering states similar to past experiences. 
As a result, TAMFormer demonstrates better performance in interactive environments and the goal association function and replan algorithm in TAMFormer further enhance its robustness in handling permutations.

\noindent \textbf{Visualization of TAM} We apply the standard t-SNE to the learned memory structure to visualize the embedding in a two-dimensional with points being colored based on groud-truth environment location. 
The visualization of the memory retrieve demonstrates that similar locations tend to cluster together, indicated by the darkness of the green color. Interestingly, the learned representation does not appear to directly reflect the semantic concept of the scene, as shown by the proximity of activated nodes related to \textit{Bowl}, which are closer to each other, while nodes within the kitchen environment are scattered around. 
We suspect that the learned representation contains internal information regarding the similarity between nodes rather than a direct mapping of semantic concepts such as \textit{kitchen}, which may be difficult to capture in a two-dimensional space.

\begin{table}[t]
    \caption{\textbf{Results of Ablation Study with Vis-Inter-Attack.} Compared with Vis-interact, the attack make it harder for model to generate the correct action without verifying the correctness of the action execution.}
   \setlength{\tabcolsep}{0.6mm}{ \begin{tabular*}{\linewidth}{@{\extracolsep{\fill}}r|cccc}
        \toprule
     Methods          & LCS$\uparrow$   & F1$\uparrow$  & F1-state$\uparrow$ & F1-Relation$\uparrow$ \\ \midrule
\rowcolor{Gray}     TAMFormer         & 64\% &  0.40 &  0.42  &  0.39 \\
     w/o replan        & 37\% &  0.20&  0.18  &   0.21    \\
     pixel-localize    & 24\% &  0.11&  0.09  &   0.13    \\
     w/o trans         & 15\% &  0.018 &  0.02  &   0.014     \\
     naive goal        & 2.7\% &  0.01 & 0.01 & 0.01    \\
        \bottomrule
    \end{tabular*}
    \vspace{-2mm}
    \label{table:ab}}
\end{table}
\begin{figure*}[ht!]
\centering
    \includegraphics[width=0.8\textwidth]{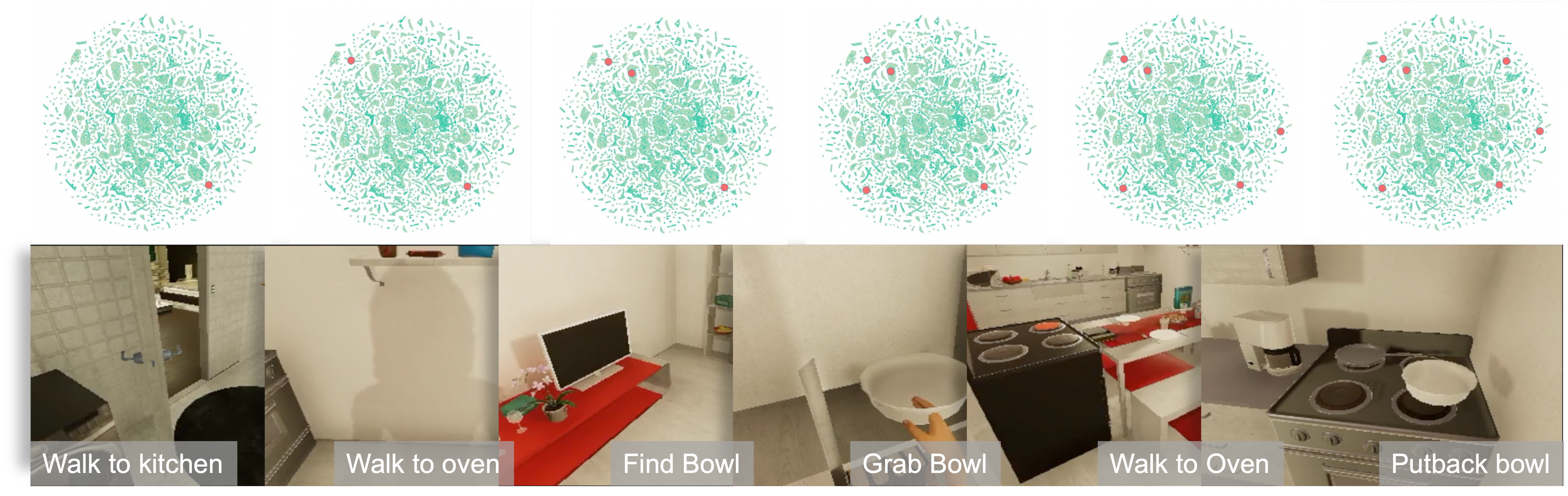}
    \caption{\textbf{Visualization of memory retrieve:} We used t-SNE on the learned memory and colorized them based on the environment location. The red dot represents the activated memory location corresponding to the current observation.}
    \label{fig:plan}
    \vspace{-3mm}
\end{figure*}

\subsection{Ablation studies}
We performed the following experiments to further investigate the proposed modules' importance.

\noindent \textbf{w/o replan:} We drop the replay algorithm during the training and planning stage, meaning we do not verify whether the current executed action is aligned with the given goal and will accept the k nearest neighbor returned from the localization network as the retrieved memory for subsequent action generation.

\noindent \textbf{pixel-localize:} Instead of training a localization network that measures the distance between two images in the latent space, inspired by~\cite{SavinovDosovitskiyKoltun2018_SPTM}, we downsampled images to the resolution of 40×30, converted it to grayscale, and used cosine distances as the difference measurement.

\noindent \textbf{w/o trans:} After retrieving the memory, we replaced the Transformer decoder as a classifier layer, meaning we only focused on the short-term actions prediction w/o considering the trajectory as a whole. This way, the action prediction will rely solely on the learned affordance representation.

\noindent \textbf{naive goal:} The goal association function is trained with the additional goal as input, making it a conditional discriminator. One variant is to remove the dependence on the goal and directly output the probability of whether the two nodes are under the same goal.

We conducted ablation experiments with two evaluation settings: vis-interact and vis-inter-att. Table \ref{table:ab} shows the performance of the full model for reference, as well as the results of each variation we tested.
With vis-interact evaluation, we found that removing the replanning module had a significant negative impact on performance, as the model could not verify if the plan had already deviated. Nevertheless, the model still outperformed the baseline method. 
Removing the learned localization network also hurt performance, as the model had to rely solely on the Transformer decoder for planning. 
We believe that using pixel differences to measure the distance between raw images could introduce too much noise, making it difficult to distinguish between similar-looking locations. Replacing the Transformer decoder with a classifier also decreased performance, as the model became short-sighted and focused only on generating the best action aligned with the retrieved memory. 
This led to repeated actions in some situations, such as when \textit{pick up} or \textit{put down} objects.
Removing the goal embedding caused the training process to hardly converge, as it was challenging to learn useful metric functions in the face of ambiguous learning scenarios.

In the vis-inter-att evaluation, we found that the w/o replan variation performed poorly, as the model was unable to recover from any mistakes made by itself or our deliberate attacks. 
Interestingly, we observed that the w/o trans variatio was less affected by attacks compared to the others. 
We suspect that short-term action prediction is more robust under attacks, as it relies less on sequential information.

    \begin{table}[ht]
        \caption{\textbf{Results of Ablation Study with Vis-Interact.} The performance decrease significantly when using naive goal association or w/o long-term sequence modeling, which shows the importance of learning the action as a whole.}
           \setlength{\tabcolsep}{0.6mm}{ \begin{tabular*}{\linewidth}{@{\extracolsep{\fill}}r|cccc}
            \toprule
        Methods          & LCS$\uparrow$   & F1$\uparrow$  & F1-state$\uparrow$ & F1-Relation$\uparrow$ \\ \midrule
\rowcolor{Gray}        TAMFormer         & 77\% &   0.46 &  0.48  &   0.44 \\ 
        w/o replan        & 63\% &  0.22&	0.18 &  0.24      \\
        pixel-localize    & 33\% &  0.12 &  0.11  &   0.12    \\
        w/o trans         & 21\% &   0.09 &  0.13  &   0.07      \\
        naive goal        & 5.8\% &  0.033&  0.04  &   0.03    \\
    
            \bottomrule
        \end{tabular*}}
        \vspace{-2mm}
        \label{table:ab}
    \end{table}

\section{Conclusion, application and limitation}

This paper proposes a novel memory model for robust visual task planning that associates human demonstration in the latent space. Specifically, we introduce a topological affordance memory that learns to extract environment information by identifying what intersections can be afforded under the current situation. 
To retrieve memories from TAM, we leverage self-supervised learning to train a localization function. 
The replanning algorithm, combined with the selective memory, enables the action generator to plan more goal-directed actions. Experimental results on an interactive simulation environment demonstrate the effectiveness of our approach in learning a meaningful representation for planning, which is robust to deviations. 
However, a major limitation of our method is the requirement for successful experiences, assuming expert demonstrations are executed without flaws. 
In real-life instructional videos, people tend to perform differently, making this assumption unreliable. To address this limitation, we suggest constructing a "bad" memory that contains failed cases, which can be used to avoid undesirable actions.

{\small
\bibliographystyle{ieee_fullname}
\bibliography{egbib}
}

\end{document}